\documentclass[11pt]{article}

\usepackage[margin=1in]{geometry}

\usepackage[T1]{fontenc}
\usepackage[utf8]{inputenc}
\usepackage[french]{babel}

\usepackage{float}
\usepackage{booktabs}

\usepackage{amsmath}
\usepackage{amssymb}

\usepackage{graphicx}

\usepackage[ruled,vlined,linesnumbered]{algorithm2e}

\usepackage{tikz}
\usetikzlibrary{arrows.meta, positioning, shapes.geometric, calc, fit}

\usepackage{xcolor}

\usepackage{hyperref}

\title{
    \textbf{Digital Twin–Supervised Reinforcement Learning Framework for Autonomous Underwater Navigation}
    
}

\author{
    Zamirddine Mari$^{1}$,
    Mohamad Motasem Nawaf $^{2}$,
    Pierre Drap$^{2}$ \\
    \\
    \small $^{1}$DGA Techniques Navales - Direction Générale de l'Armement, Toulon, France \\
    \small $^{2}$LIS, CNRS, Aix-Marseille University, Marseille, France \\
    \small Corresponding author: zamirddine.mari@intradef.gouv.fr
}

\date{} 

\begin{document}

\maketitle

\begin{center}
\textit{This document is a preprint prepared for submission to Sensors (MDPI).\\
The title and content are preliminary and may be updated in future versions.}
\end{center}
\bigskip

\renewcommand{\abstractname}{Abstract}
\begin{abstract}
Autonomous navigation in underwater environments remains a major challenge due to the absence of GPS, degraded visibility, and the presence of submerged obstacles. This article investigates these issues through the case of the \textit{BlueROV2}, an open platform widely used for scientific experimentation. We propose a deep reinforcement learning approach based on the Proximal Policy Optimization (PPO) algorithm, using an observation space that combines target-oriented navigation information, a virtual occupancy grid, and ray-casting along the boundaries of the operational area. The learned policy is compared against a reference deterministic kinematic planner, the \textit{Dynamic Window Approach} (DWA), commonly employed as a robust baseline for obstacle avoidance. The evaluation is conducted in a realistic simulation environment and complemented by validation on a physical \textit{BlueROV2} supervised by a 3D digital twin of the test site, helping to reduce risks associated with real-world experimentation. The results show that the PPO policy consistently outperforms DWA in highly cluttered environments, notably thanks to better local adaptation and reduced collisions. Finally, the experiments demonstrate the transferability of the learned behavior from simulation to the real world, confirming the relevance of deep RL for autonomous navigation in underwater robotics.
\end{abstract}

\bigskip

\section{Introduction}

The autonomy of underwater vehicles has become a major challenge for the exploration, monitoring, and inspection of marine environments. Underwater robots—whether fully autonomous AUVs or semi-autonomous ROVs—play an increasingly important role in scientific observation, bathymetric mapping, offshore infrastructure inspection, maintenance operations, and security or defense missions. The ability of these vehicles to navigate reliably and safely in complex, deep, or remote environments directly affects the quality, cost, and efficiency of such operations. In this context, improving autonomous navigation capabilities is essential to reducing dependence on teleoperation, extending mission duration, and increasing operational safety.

However, underwater environments impose severe constraints on autonomous navigation. The absence of GPS, heavily degraded visibility, hydrodynamic disturbances, and the presence of static obstacles (terrain features, rocks, wrecks, infrastructure) and dynamic obstacles (marine life, current variations) make trajectory planning difficult and uncertain. Moreover, the diversity of operational contexts—from cluttered coastal zones to industrial structures and complex natural environments—requires real-time adaptation capabilities. Ensuring safe behaviours therefore demands strategies capable of handling obstacle avoidance, stability control, and continuous local situational awareness. Finally, real-world experimental validation is costly and risky, which reinforces the importance of advanced simulation and digital twins for training and evaluating autonomous systems under realistic yet controlled conditions.

A robust way to tackle these challenges is to rely on open and modular experimental platforms that allow researchers to explore, test, and compare different autonomous navigation strategies in constrained environments. Among these platforms, the \textit{BlueROV2} has emerged as a reference vehicle for scientific research and algorithm development. Its open architecture, relatively low cost, and extensible software ecosystem make it an ideal support for the study of advanced navigation techniques, particularly those based on reinforcement learning. In this perspective, recent works have focused on three main axes: (i) navigation and obstacle avoidance approaches based on RL; (ii) research specifically involving the \textit{BlueROV2} as an experimental platform; and (iii) contributions leveraging digital twins to bridge simulation and reality in complex underwater environments.

\subsection{Autonomous navigation and obstacle avoidance using reinforcement learning}

The application of reinforcement learning (RL) to underwater navigation is an active and rapidly growing research area. Bhopale \textit{et al.}~\cite{ref_1} propose an RL-based obstacle avoidance technique enabling an AUV to learn effective behaviours without requiring a detailed dynamic model. Eweda and ElNaggar~\cite{ref_2} provide an extensive review of the challenges faced by AUVs in dynamic environments, highlighting the ability of deep RL to produce robust policies despite disturbances and energy constraints.

Several methodological developments have since emerged to improve the stability and efficiency of learning. Marchel \textit{et al.}~\cite{ref_3} demonstrate the benefits of \textit{Curriculum Learning} for training an RL agent to navigate environments of progressively increasing difficulty. Liu \textit{et al.}~\cite{ref_4} explore offline RL for obstacle avoidance in an underactuated AUV, while Manderson \textit{et al.}~\cite{ref_5} leverage a goal-conditioned architecture to learn visual behaviours directly from images.

These contributions reflect a growing interest in RL as a solution for advanced autonomy. However, to rigorously evaluate its real-world potential, it is essential to compare RL-based methods with established deterministic baselines.

\subsection{Comparison between deterministic approaches and learning-based methods: DWA as a baseline against RL}

In the mobile robotics literature, the \textit{Dynamic Window Approach} (DWA)~\cite{ref_6} remains one of the most widely used reactive obstacle avoidance algorithms. Its simplicity, computational efficiency, and native integration into ROS make it a standard baseline for assessing the advantages of RL policies.

Several studies have undertaken direct comparisons between RL and DWA. Patel \textit{et al.}~\cite{ref_7} show that DRL policies can outperform DWA in environments populated with moving obstacles, notably by reducing collisions and dynamic constraint violations. Arce \textit{et al.}~\cite{ref_8} conduct a structured comparative evaluation including DWA, TEB, CADRL, and a SAC agent, observing that RL agents generally surpass deterministic methods in complex environments. Yeom \textit{et al.}~\cite{ref_9} demonstrate that a DRL controller produces more efficient and smoother trajectories than DWA in a wheeled-ground-robot scenario.

These comparative studies indicate that RL approaches tend to offer superior performance in dynamic, congested, or unstructured environments—an especially valuable property for underwater systems operating in unpredictable conditions.

\subsection{Research on the \textit{BlueROV2} platform}

In parallel with methodological advances, several works specifically focus on the \textit{BlueROV2} platform. Wilby \textit{et al.}~\cite{ref_10} introduce a modified open-source variant called \textit{Makobot}, optimised for autonomous missions. Willners~\cite{ref_11} demonstrates the feasibility of transforming commercial ROVs such as the \textit{BlueROV2} into low-cost AUVs by adding onboard navigation and control capabilities.

However, despite its growing popularity as a research platform, few studies investigate the use of RL for obstacle avoidance on the \textit{BlueROV2}. This gap highlights the relevance of developing and evaluating learning-based policies on this accessible and widely adopted platform.

\subsection{Digital twins and underwater modelling}

Recent advances in simulation and digital twins have also strengthened validation capabilities in underwater robotics. Scaradozzi \textit{et al.}~\cite{ref_12} and Lambertini \textit{et al.}~\cite{ref_13} show that digital twin architectures enable fine-grained modelling of the robot and its environment, facilitating supervision and planning. Adetunji \textit{et al.}~\cite{ref_14} demonstrate that such systems improve teleoperation in complex scenarios.

These hybrid simulation–reality approaches are essential tools for training and testing RL policies prior to real-world deployment, helping to reduce risks and costs.

\subsection{Summary and positioning}

In summary, the literature demonstrates the relevance of RL for underwater obstacle avoidance while emphasising the importance of comparing it with established deterministic methods such as DWA to rigorously assess its benefits. At the same time, the \textit{BlueROV2} has become a preferred experimental platform, and digital twins now offer powerful tools to secure and accelerate the transition from simulation to reality.

The present work lies at the intersection of these three research directions. It offers a controlled validation of using a PPO policy for autonomous navigation of a \textit{BlueROV2}, leveraging a realistic 3D environment capable of accurately simulating obstacles and local interactions. This approach enables the safe and reproducible evaluation of an RL agent’s ability to surpass a robust deterministic baseline such as DWA.

This positioning represents a significant step: it establishes the feasibility of RL-based autonomous control on a real underwater vehicle and demonstrates the value of coupling simulation with a physical robot. Building on this foundation, future work will extend these contributions toward integrating real perception sensors (video, sonar) and conducting trials in underwater environments featuring authentic obstacles and more diverse conditions. The long-term objective is to advance toward fully autonomous, robust, and deployable real-world underwater navigation.

\section{Materials and Methods}

\begin{figure}[h!]
\centering
\includegraphics[width=0.8\linewidth]{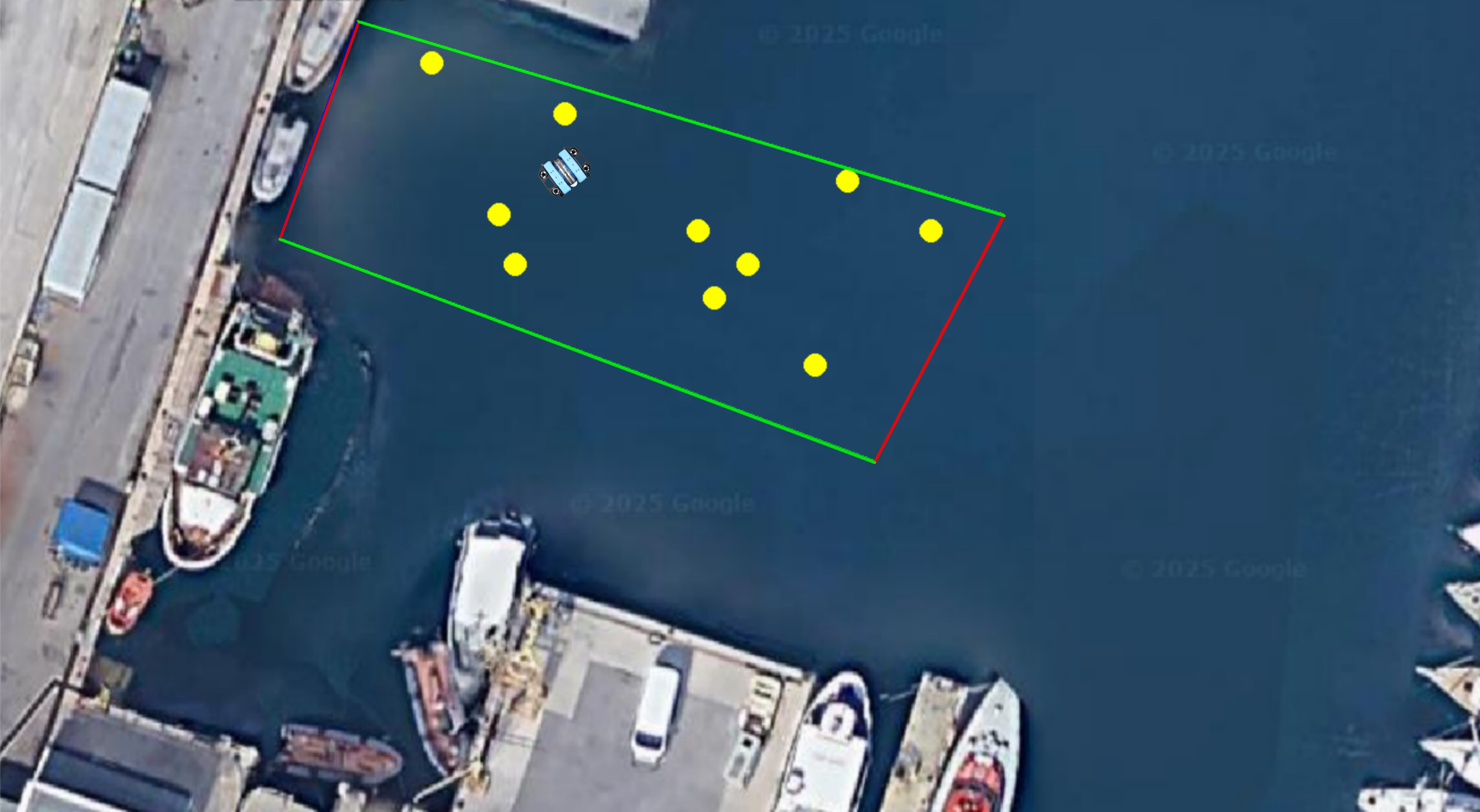}
\caption{Screenshot of the 2D Python visualisation tool showing the random placement of obstacles within the rectangular area used for both RL and DWA tests.}
\label{fig:pygame}
\end{figure}

This section describes the methodological approach adopted to study the autonomous navigation of the \textit{BlueROV2} underwater vehicle in a partially unknown environment containing submerged obstacles. Concretely, the navigation scenario illustrated in Figure~\ref{fig:pygame} consists in moving the vehicle at a fixed depth from one side of a rectangular area to the other while avoiding approximately ten static obstacles.

The objective is to evaluate the ability of a deep reinforcement learning approach to ensure safe and efficient navigation in direct comparison with a classical motion planner. Two paradigms are therefore considered: (i) a deterministic method based on the \textit{Dynamic Window Approach} (DWA), and (ii) a learning-based method using \textit{Reinforcement Learning} (RL) with the \textit{Proximal Policy Optimization} (PPO) algorithm~\cite{ref_15}.

The first subsection introduces the comparative framework between these two approaches, while the second formalises the problem as a Markov Decision Process (MDP), which serves as the basis for training the RL policy.

\subsection{Comparative Framework of the Approaches}

We begin by establishing the comparative framework between the two studied methods by first presenting the theoretical foundations and implementation principles of the \textit{Dynamic Window Approach} (DWA). The reinforcement learning approach, based on the PPO algorithm, is then introduced and detailed in the following section.

\begin{enumerate}
    \item \textbf{Deterministic Approach: Dynamic Window Approach (DWA)}  

    We provide here a description of how the \textit{Dynamic Window Approach} algorithm was implemented. DWA relies on a predictive evaluation of a discrete set of admissible kinematic commands for the vehicle. At each cycle, the algorithm considers a family of actions parameterised by an angular variation and a traversable distance, defining a finite set of possible future configurations.

    \medskip
    \textbf{Predictive Model.}
    For each candidate action, the vehicle projects its future position:
    \[
        \mathbf{x}' = \mathbf{x} + d \,
        \begin{pmatrix}
            \cos(\theta + \Delta\theta) \\
            \sin(\theta + \Delta\theta)
        \end{pmatrix},
    \]
    where $\mathbf{x}$ is the current position, $\theta$ the orientation, $\Delta\theta$ the candidate angular variation, and $d$ the forward distance.

    \medskip
    \textbf{Distance to the Goal.}
    The attraction cost toward the goal, defined as the geometric centre of the exit gate, is:
    \[
        C_{\text{goal}}(\mathbf{x}') =
        \left\| \mathbf{x}' - \mathbf{g} \right\|,
    \]
    where $\mathbf{g}$ denotes the target point.

    \medskip
    \textbf{Obstacle Clearance.}
    For each predicted configuration, the minimum safety distance to obstacles is computed as:
    \[
        \delta(\mathbf{x}') =
        \min_{i} \left(
        \lVert \mathbf{x}' - \mathbf{o}_i \rVert
        - r_i - r_{\mathrm{robot}} - m_{\mathrm{s}}
        \right),
    \]
    where $\mathbf{o}_i$ and $r_i$ are respectively the position and radius of obstacle $i$, $r_{\mathrm{robot}}$ is the robot’s safety radius, and $m_{\mathrm{s}}$ is an additional safety margin.  
    A configuration is rejected if $\delta(\mathbf{x}') \leq 0$.  
    A normalised clearance score is defined as:
    \[
        S_{\text{clear}}(\mathbf{x}')
        = \min\!\left( 1, \frac{\delta(\mathbf{x}')}{D_{\max}} \right).
    \]

    \medskip
    \textbf{Kinematic Progress.}
    The forward progress is represented by the travelled distance:
    \[
        S_{\text{prog}} = d.
    \]

    \medskip
    \textbf{Global Cost Function.}
    Each command is evaluated using a weighted combination:
    \[
        J(\mathbf{x}') =
        - \alpha \, C_{\text{goal}}(\mathbf{x}')
        + \beta \, S_{\text{clear}}(\mathbf{x}')
        + \gamma \, S_{\text{prog}},
    \]
    where $(\alpha,\beta,\gamma)$ modulate the relative influence of each criterion.  
    The selected action is then:
    \[
        (\Delta\theta^\star, d^\star)
        = \arg\max_{\Delta\theta, d} J(\mathbf{x}').
    \]

    This fully reactive method is a standard in mobile robotics. However, its effectiveness strongly depends on parameter tuning and the accuracy of sensed data, which may limit performance in dense, noisy, or structurally complex environments.

    \item \textbf{Learning-Based Approach: Reinforcement Learning (RL)}  

    The proposed learning-based method relies on an agent trained using the PPO algorithm. Unlike DWA, RL does not explicitly rely on a kinematic model: it constructs an optimal policy through interaction with the environment based on received rewards. This model-free nature promotes robustness, adaptability, and the ability to generalise in unpredictable environments.
\end{enumerate}

The comparison between these two paradigms aims to determine to what extent a learned policy can rival a classical deterministic planner. The next section is therefore devoted to establishing the theoretical framework and implementation details of the reinforcement learning approach.

\subsection{Reinforcement Learning Problem Formulation}

The objective is to determine an optimal control policy that guides the vehicle toward its target while ensuring safe navigation.

This problem is formulated as a \textit{Markov Decision Process} (MDP)~\cite{ref_16}, defined by the tuple $\langle \mathcal{S}, \mathcal{A}, \mathcal{P}, \mathcal{R} \rangle$, where:  
\begin{itemize}
    \item $\mathcal{S}$ is the set of observable system states (robot position, heading, obstacle distances, etc.);
    \item $\mathcal{A}$ is the set of admissible actions, here corresponding to linear and angular velocity commands;
    \item $\mathcal{P}(s'|s,a)$ is the transition probability from state $s$ to state $s'$ after executing action $a$;
    \item $\mathcal{R}(s,a)$ is the immediate reward measuring the quality of the action (progress toward the target, obstacle avoidance, etc.).
\end{itemize}

The goal of the agent is to maximise the cumulative return defined as:
\begin{equation}
    G_t = \sum_{k=0}^{\infty} \gamma^k R_{t+k+1},
\end{equation}
where $0 \leq \gamma < 1$ is the discount factor weighting future rewards. The agent therefore learns a stochastic policy $\pi(a|s)$ that maximises the expected return $\mathbb{E}[G_t]$.

To solve this MDP, we use a deep reinforcement learning method based on the \textit{Proximal Policy Optimization} (PPO) algorithm. PPO is a policy optimisation method that improves training stability by constraining successive policy updates through a clipped objective:
\begin{equation}
    L^{\mathrm{CLIP}}(\theta) = 
    \mathbb{E}_t \left[ 
        \min \left( 
            r_t(\theta) \hat{A}_t, 
            \mathrm{clip}(r_t(\theta), 1 - \epsilon, 1 + \epsilon) \hat{A}_t
        \right) 
    \right],
\end{equation}
where $r_t(\theta) = \frac{\pi_\theta(a_t|s_t)}{\pi_{\theta_{\text{old}}}(a_t|s_t)}$ is the probability ratio, $\hat{A}_t$ an estimate of the advantage, and $\epsilon$ a trust-region hyperparameter typically between 0.1 and 0.2. PPO is chosen for its robustness, ease of implementation, and strong performance in similar continuous environments~\cite{ref_15}.

\subsection{Simulation Environment}

\begin{figure}[H]
    \centering
    \includegraphics[width=0.8\linewidth]{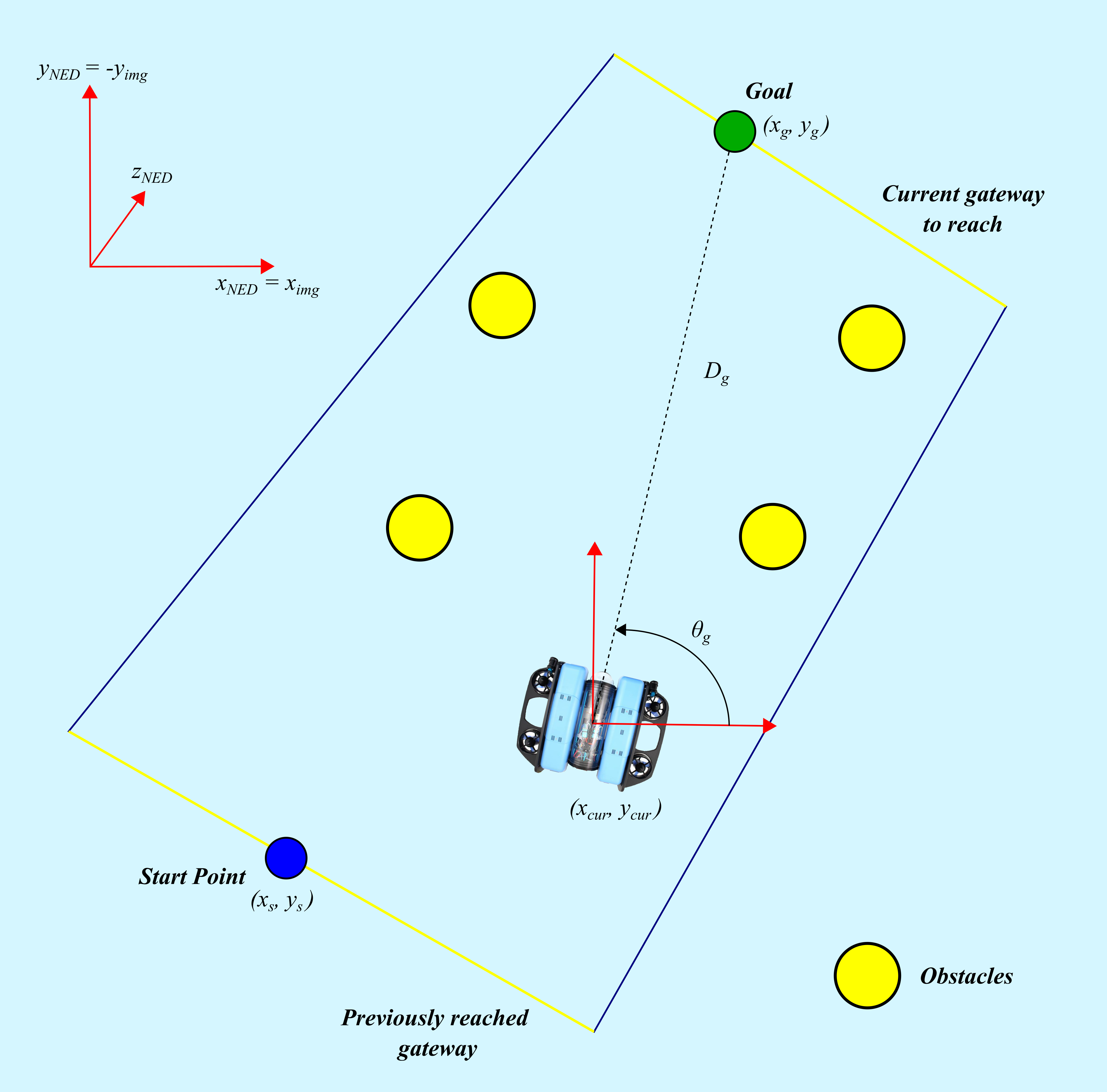}
    \caption{Diagram illustrating the interaction between the navigation agent and the simulated environment.}
    \label{fig:env_interaction}
\end{figure}

The simulation environment serves as the training and evaluation framework for the \textit{BlueROV2} agent. The goal is to guide the vehicle from an initial position located on one side of a quadrilateral representing the workspace to a target position placed on the opposite side. This space contains submerged obstacles whose positions are not known \textit{a priori} by the agent. The control policy must therefore steer the vehicle to the target while respecting spatial constraints and avoiding collisions.

\subsubsection{Global Frame Modelling and Coordinate Transformation}

The environment is described in the conventional marine robotics coordinate system \textit{North–East–Down (NED)}:
\begin{equation}
\begin{cases}
x_{\text{NED}}(t): \text{North coordinate}, \\
y_{\text{NED}}(t): \text{East coordinate}, \\
z_{\text{NED}}(t): \text{Down coordinate}.
\end{cases}
\end{equation}

For horizontal-plane trajectory planning, only $(x_{\text{NED}}, y_{\text{NED}})$ are considered. The graphical representation and numerical simulation use an image frame $(x_{\text{img}}, y_{\text{img}})$ defined as:
\begin{equation}
\begin{cases}
x_{\text{img}}(t) = y_{\text{NED}}(t), \\
y_{\text{img}}(t) = -\,x_{\text{NED}}(t).
\end{cases}
\end{equation}
Thus, motion toward the North $(+x_{\text{NED}})$ corresponds to a decrease in $y_{\text{img}}$, while motion toward the East $(+y_{\text{NED}})$ increases $x_{\text{img}}$.

\subsubsection{Kinematic State and Autonomous Navigation Objective}

The kinematic state of the \textit{BlueROV2} is defined by:
\begin{equation}
\mathbf{s}_{\text{NED}}(t) =
\begin{bmatrix}
x_{\text{NED}}(t) \\
y_{\text{NED}}(t) \\
\theta(t)
\end{bmatrix},
\end{equation}
where $(x_{\text{NED}}, y_{\text{NED}})$ is the position and $\theta(t)$ the orientation with respect to the positive East axis ($y_{\text{NED}}$, corresponding to the horizontal axis of the image frame). The planar dynamics are:
\begin{equation}
\begin{cases}
\dot{x}_{\text{NED}}(t) = v(t) \sin \theta(t), \\
\dot{y}_{\text{NED}}(t) = v(t) \cos \theta(t),
\end{cases}
\end{equation}
where $v(t)$ is the translational velocity. Autonomous navigation aims to generate a trajectory $\Gamma(t)$ from the initial state $\mathbf{s}(t_0)$ to the final state $\mathbf{s}(t_f)$, while ensuring:
\begin{equation}
\Gamma(t) \subset \mathcal{Q} \setminus \bigcup_{i=1}^n O_i, \quad \forall t \in [t_0, t_f],
\end{equation}
where $\mathcal{Q}$ is the working quadrilateral and $\{O_1, \dots, O_n\}$ the set of submerged obstacles.

\subsection{Observation Space of the Autonomous Navigation Agent}

The observation space defines all variables perceived by the agent at each navigation step. In our case, the agent receives an observation vector that integrates both target-oriented navigation information and local obstacle detection.

\subsubsection{Distance and Direction to the Goal}

Let $(x(t),y(t))$ be the vehicle position and $(x_g, y_g)$ the target (gateway) centre. The normalised distance to the goal is:
\begin{equation}
    d_g(t) = \frac{\| (x(t), y(t)) - (x_g, y_g) \|}{d_{\max}},
\end{equation}
with $d_{\max}$ a normalisation constant corresponding to the maximum possible distance. The relative angle between the vehicle heading $\theta(t)$ and the direction to the goal is:
\begin{equation}
    \Delta \theta_g(t) = \arctan2(y_g - y(t), x_g - x(t)) - \theta(t),
\end{equation}
mapped to $[-\pi, \pi]$ and normalised to $[0,1]$:
\begin{equation}
    \tilde{\theta}_g(t) = \frac{|\Delta \theta_g(t)|}{\pi}.
\end{equation}

Thus, the first observation component is:
\begin{equation}
    o_1(t) = 
    \begin{bmatrix}
        d_g(t) \\
        \tilde{\theta}_g(t)
    \end{bmatrix}.
\end{equation}

\subsubsection{Obstacle Detection via Occupancy Grid}

\begin{figure}[H]
    \centering
    \includegraphics[width=0.8\linewidth]{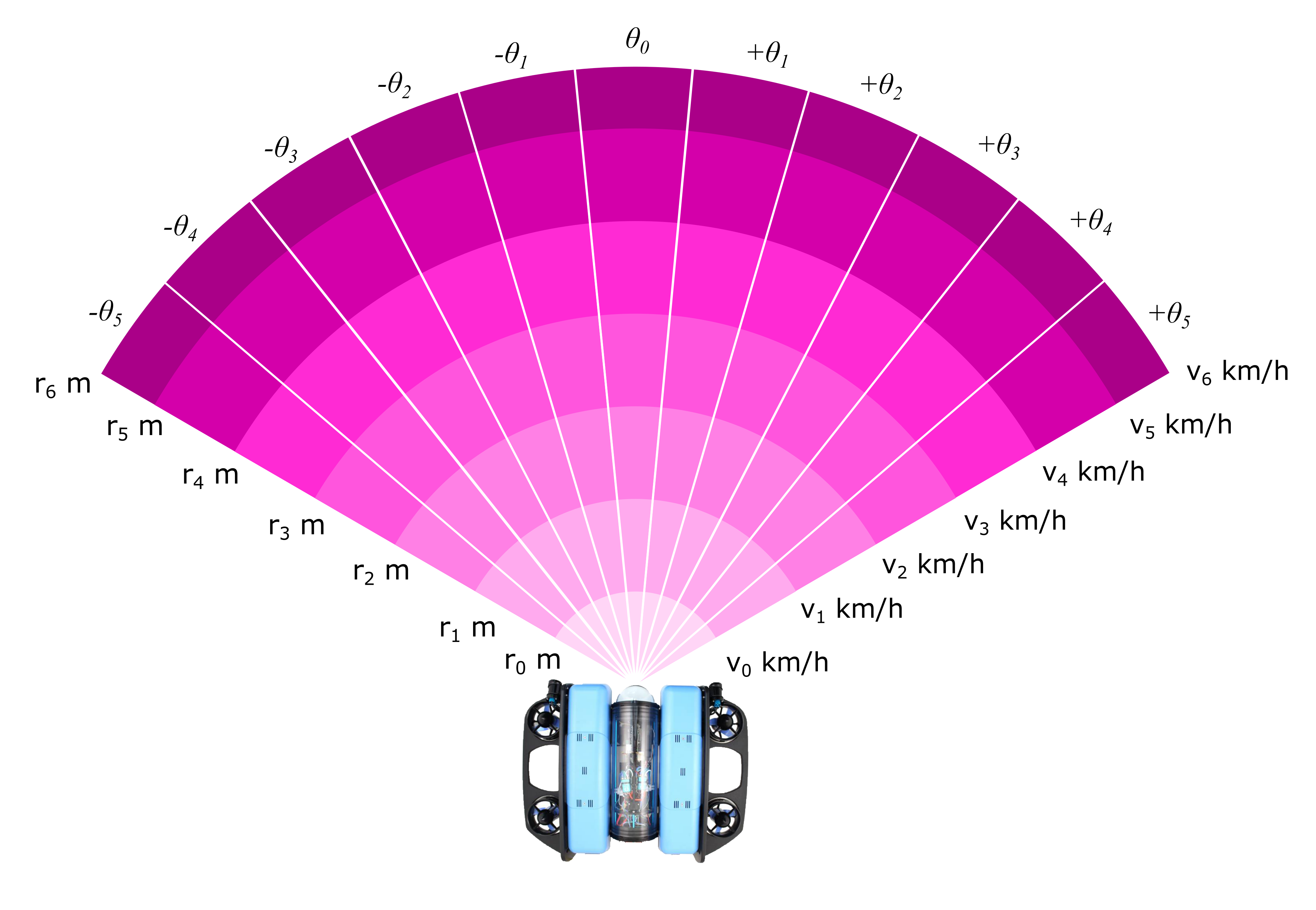}
    \caption{Virtual occupancy grid used for obstacle detection.}
    \label{fig:occ_grid}
\end{figure}

Obstacle perception is based on a virtual occupancy grid inspired by a forward-looking sonar. We consider discrete velocity values $\mathcal{V}$, detection distances $\mathcal{D}$, and angular directions $\mathcal{A}$. Each triplet $(v_i, d_j, \alpha_k)$ defines a grid cell indicating whether a trajectory following this configuration would lead to a collision. The occupancy indicator is:
\begin{equation}
    O_{j,k} =
    \begin{cases}
        1 & \text{if an obstacle is detected in sector } (d_j, \alpha_k), \\
        0 & \text{otherwise}.
    \end{cases}
\end{equation}
The associated observation vector is:
\begin{equation}
    o_2(t) = \big[ O_{j,k}(t) \big]_{j=1..n,\,k=1..p}.
\end{equation}

\subsubsection{Raycasting for Workspace Boundary Awareness}

\begin{figure}[H]
    \centering
    \includegraphics[width=0.8\linewidth]{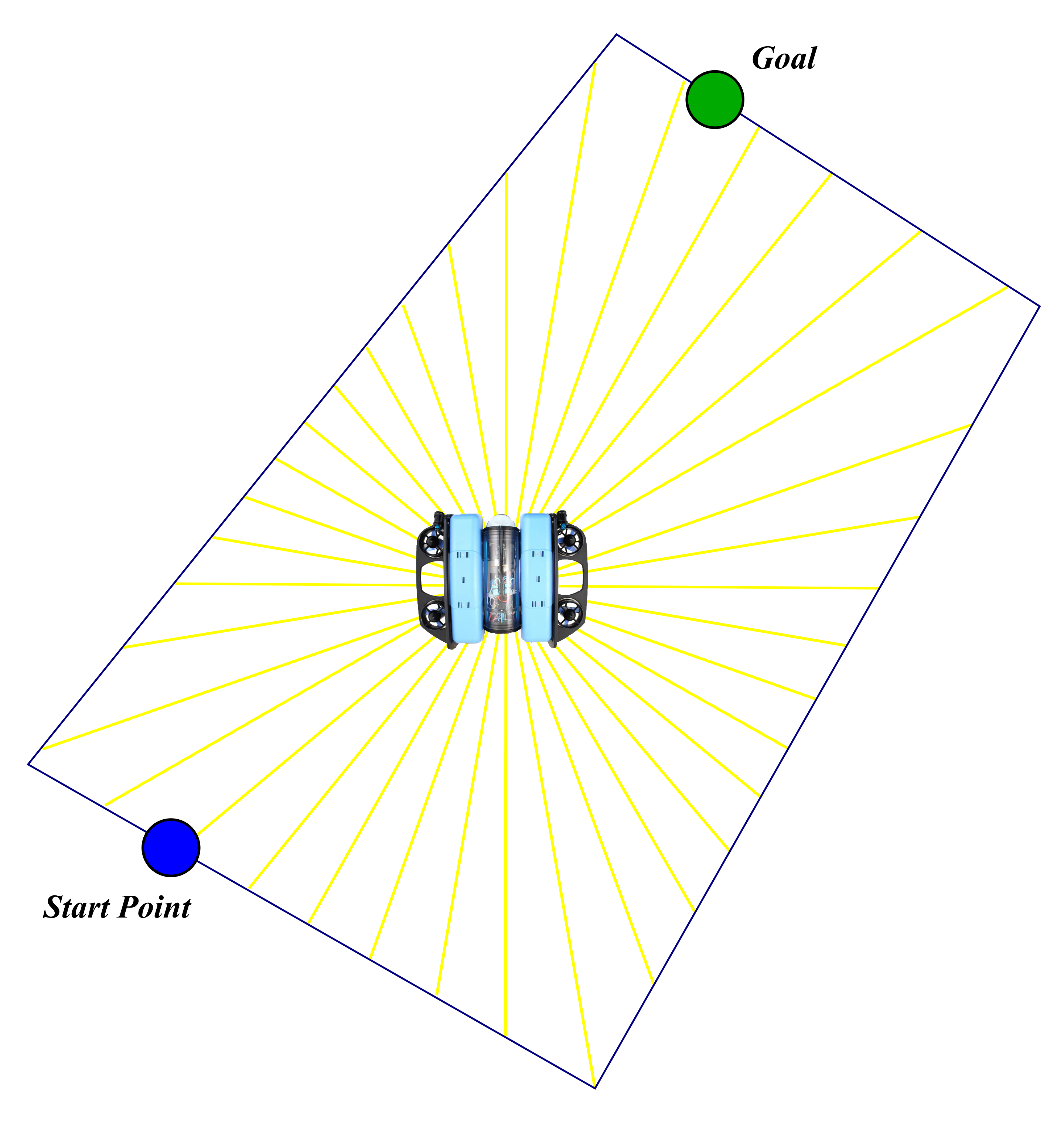}
    \caption{Virtual rays used to ensure that the vehicle remains inside the workspace.}
    \label{fig:raycasting}
\end{figure}

To ensure that the agent remains within the quadrilateral workspace, a set of rays $\{r_1, \dots, r_q\}$ is cast from the vehicle in predefined angular directions. The normalised length of each ray is:
\begin{equation}
    \rho_\ell(t) = \frac{\ell_{\min}(r_\ell)}{\ell_{\max}},
\end{equation}
where $\ell_{\min}(r_\ell)$ is the distance from the ray origin to the first intersection with the quadrilateral and $\ell_{\max}$ the maximum ray length. The resulting observation vector is:
\begin{equation}
    o_3(t) = \big[ \rho_\ell(t) \big]_{\ell=1..q}.
\end{equation}

\subsubsection{Final Observation Vector}

The full observation vector is the concatenation:
\begin{equation}
    o(t) = \big[ o_1(t), o_2(t), o_3(t) \big].
\end{equation}
This observation space provides the agent with all necessary information to navigate toward the target while avoiding obstacles and respecting workspace constraints.

\subsection{Action Space of the Autonomous Navigation Agent}

\begin{figure}[H]
    \centering
    \includegraphics[width=0.8\linewidth]{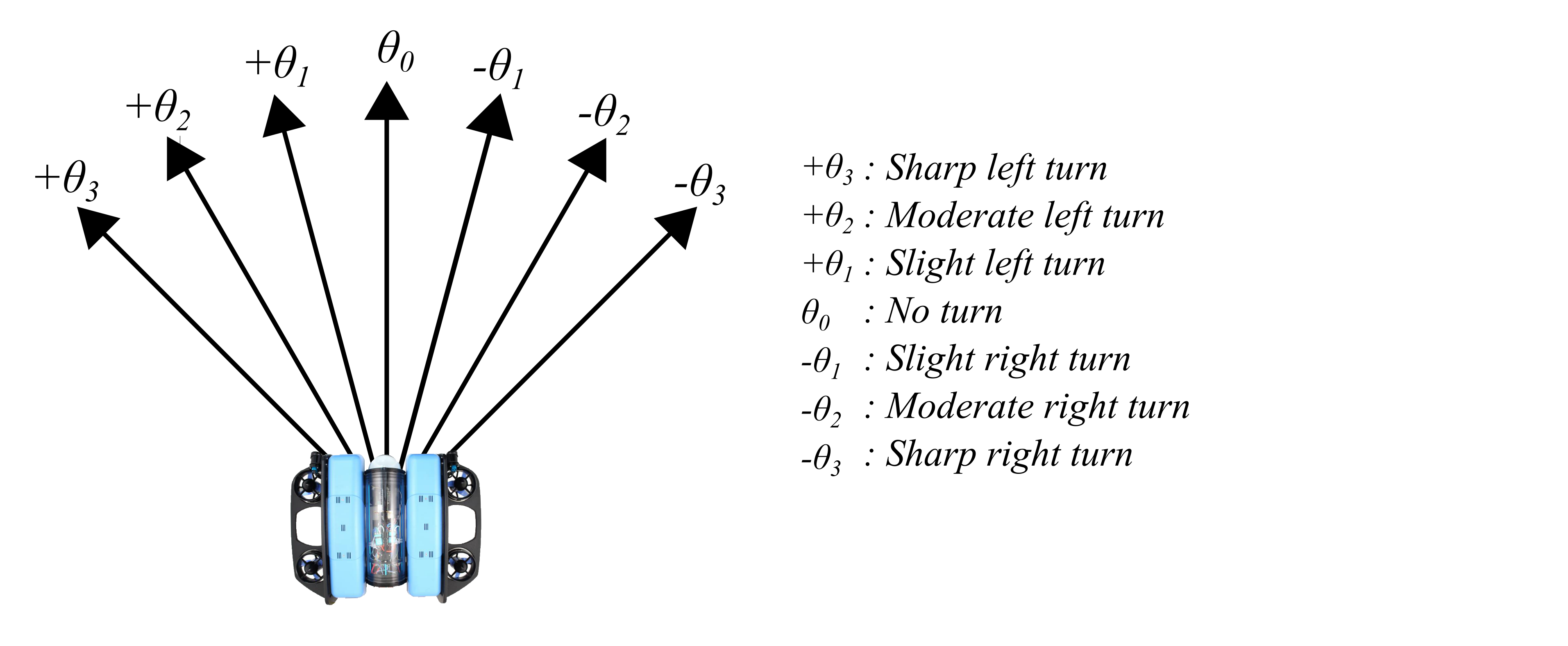}
    \caption{Discretisation of actions into elementary angular variations.}
    \label{fig:actions}
\end{figure}

The action space corresponds to all elementary decisions that the agent can take to modify its trajectory. For the \textit{BlueROV2}, the action selected at each step consists of a discrete change in orientation applied to the vehicle’s current heading.

We define a finite set of $N=7$ actions corresponding to discrete angular variations in the interval $[-\pi/4, +\pi/4]$. Each action $a \in \mathcal{A}$ is associated with:
\begin{equation}
    \mathcal{A} = \{ a_0, a_1, \dots, a_6 \}, \qquad
    \Delta \theta_a \in \left\{ -\frac{\pi}{4}, -\frac{\pi}{6}, -\frac{\pi}{12}, 0, \frac{\pi}{12}, \frac{\pi}{6}, \frac{\pi}{4} \right\}.
\end{equation}
Thus, $a_0$ corresponds to a sharp left turn ($-45^\circ$), while $a_6$ corresponds to a sharp right turn ($+45^\circ$). The central action $a_3$ maintains the current heading.

Let the kinematic state at time $t$ be:
\begin{equation}
    s(t) = 
    \begin{bmatrix}
        x(t) \\ y(t) \\ \theta(t)
    \end{bmatrix}.
\end{equation}
When an action $a$ is applied, the orientation becomes:
\begin{equation}
    \theta(t+1) = \theta(t) + \Delta \theta_a \quad \mathrm{mod}\; 2\pi,
\end{equation}
and the position updates with constant speed $v$ and step size $\Delta d$:
\begin{equation}
    \begin{bmatrix}
        x(t+1) \\ y(t+1)
    \end{bmatrix}
    =
    \begin{bmatrix}
        x(t) + \Delta d \cos(\theta(t+1)) \\
        y(t) + \Delta d \sin(\theta(t+1))
    \end{bmatrix}.
\end{equation}

Each action represents an elementary predicted trajectory evaluated through the occupancy grid, allowing the agent to select actions that avoid obstacles while progressing toward the target.

\subsubsection{Reward Shaping}

We define a reward function composed of three elements: (i) local progress, (ii) intermediate milestones, and (iii) terminal events (success, collision, track exit).

\paragraph{1) Local Progress}

Normalised progress between entry and exit of the workspace corridor is:
\[
p_t = \mathrm{progress}(x_t) \in [0,1],
\qquad
\Delta p_t = p_t - p_{t-1}.
\]

Strictly positive progress is rewarded via:
\[
r^{\text{prog}}_t = b_{\text{prog}} \, p_t ,
\]
where \(b_{\text{prog}}\) captures coefficients related to direction and distance.  
If the agent does not progress (\(\Delta p_t \le 0\)), no reward is given and no penalty is applied in this version of the \texttt{step()} function.

\paragraph{2) Intermediate Milestones}

To reinforce long-term structure, three intermediate milestones are defined:
\[
p_t \in \{0.25,\ 0.50,\ 0.75\}.
\]
When a milestone is crossed between two time steps:
\[
r^{\text{milestone}}_t =
\begin{cases}
B_{1/4}, & \text{if } 0.25 \text{ is crossed},\\
B_{1/2}, & \text{if } 0.50 \text{ is crossed},\\
B_{3/4}, & \text{if } 0.75 \text{ is crossed}.
\end{cases}
\]
Thus, for \(\Delta p_t > 0\), the progress reward is:
\[
r^{\text{progress}}_t = r^{\text{prog}}_t + r^{\text{milestone}}_t.
\]

\paragraph{3) Obstacle Avoidance and Safety (Terminal Events)}

Two terminal negative events produce a penalty and immediately end the episode:
\[
\text{collision}(x_t) 
\quad\text{or}\quad 
\neg \text{inside\_track}(x_t)
\quad\Rightarrow\quad 
r_t = B_{\text{fail}} < 0.
\]

\paragraph{4) Terminal Success Reward}

Reaching the exit gate terminates the episode with a positive reward:
\[
r^{\text{success}} = B_{\text{succ}}.
\]

\paragraph{5) Final Instantaneous Reward}

The reward returned at each time step is:
\[
r_t =
\begin{cases}
B_{\text{fail}}, & \text{if collision or track exit},\\[3pt]
B_{\text{succ}}, & \text{if gateway reached},\\[3pt]
r^{\text{progress}}_t, & \text{if } \Delta p_t > 0,\\[3pt]
0, & \text{otherwise}.
\end{cases}
\]

\section{Experimental Results}

This section presents the results obtained during the training, evaluation, and validation of the deep reinforcement learning model developed for the autonomous navigation of the \textit{BlueROV2}. The analysis includes:
(i) a description of the training parameters,
(ii) the evolution of the agent’s training performance,
(iii) a quantitative comparison with the \textsc{DWA} algorithm,
(iv) validation on a real robot at sea.

\subsection{Training Parameters}

Below are the training parameters used to train the model with Ray version~2.49.2 and its RLlib framework dedicated to reinforcement learning.

\subsubsection{Model Architecture}

The model is based on a \textsc{PPO} architecture using an MLP network that processes an 84-dimensional state vector and includes:
\begin{itemize}
    \item three fully connected layers of size $[128,\,128,\,128]$;
    \item \texttt{tanh} activation functions for the main layers;
    \item observation normalization;
    \item separate actor–critic heads for PPO;
    \item no convolutional layers (vector-based observations only).
\end{itemize}

\subsubsection{Training Hyperparameters}

The main hyperparameters used to train the autonomous navigation agent are summarised in Table~\ref{tab:hparams_bluerov}.

\begin{table}[H]
\centering
\caption{Main hyperparameters used for PPO training of the BLUEROV2 agent.}
\label{tab:hparams_bluerov}
\begin{tabular}{ll}
\toprule
\textbf{Hyperparameters} & \textbf{Values} \\
\midrule
Learning rate & \texttt{2.5e-4} \\
Gamma ($\gamma$) & \texttt{0.95} \\
$\lambda$ (GAE) & \texttt{1.0} \\
Clip parameter & \texttt{0.3} \\
KL target & \texttt{0.01} \\
Entropy coeff & \texttt{0.0} \\
VF loss coeff & \texttt{1.0} \\
VF clip param & \texttt{10.0} \\
Num epochs & \texttt{30} \\
Rollout fragment length & \texttt{30} \\
Train batch size & \texttt{1950} \\
Minibatch size & \texttt{128} \\
\bottomrule
\end{tabular}
\end{table}

\subsubsection{Compute Resources}

Training was carried out on a Windows 11 workstation equipped with:

\begin{itemize}
    \item \textbf{CPU}: Intel Core i9–10900K (20 logical threads),
    \item \textbf{GPU}: NVIDIA GeForce RTX 3080,
    \item \textbf{RAM}: 32 GB.
\end{itemize}

\section{Analysis of Training Performance}

We analyse here the performance of the autonomous navigation agent by focusing on two key metrics: \textit{arrival\_success\_mean}, an indicator of the average success rate, and \textit{episode\_return\_mean}, representing the cumulative reward per episode.

Performance is evaluated through two main metrics:
(i) the success rate (reaching the final waypoint without collision),  
(ii) the average episode return (cumulative reward).

\begin{figure}[h!]
\centering
\includegraphics[width=0.92\linewidth]{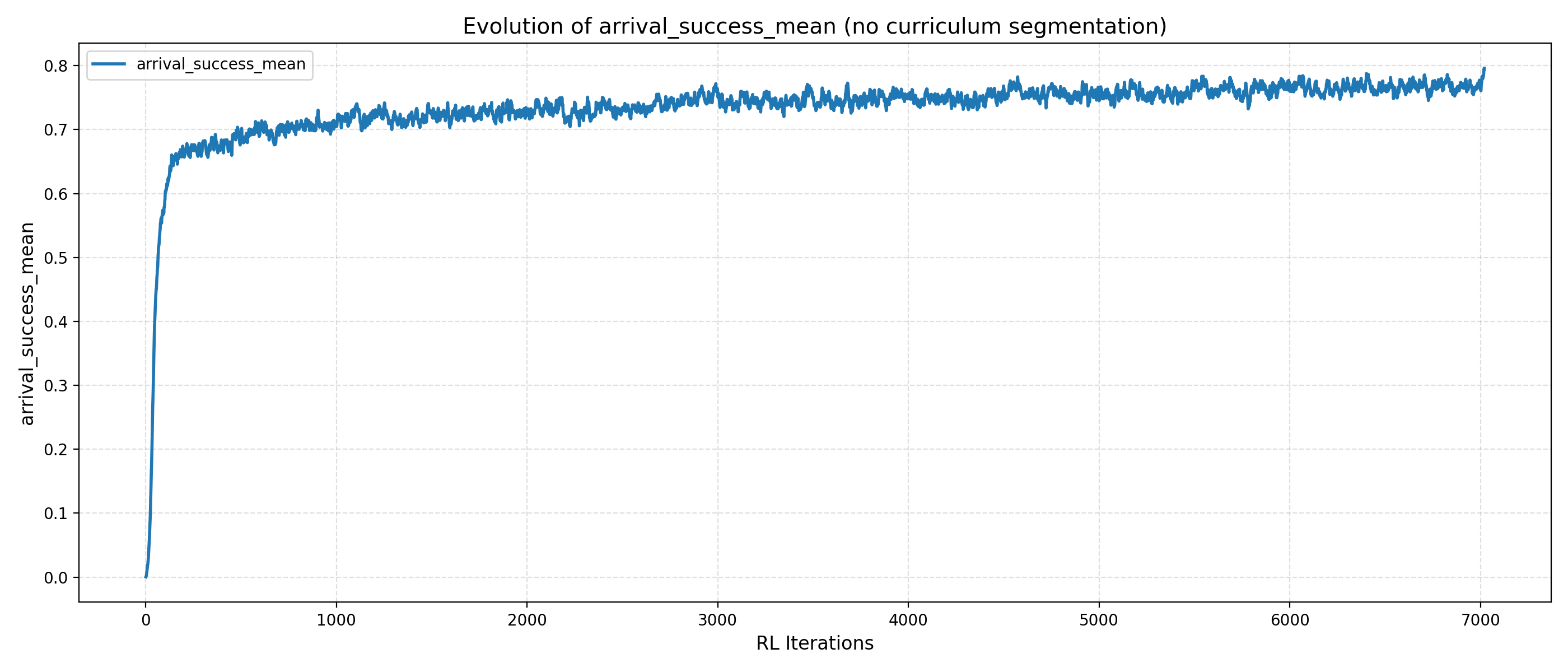}
\caption{Average success rate during BlueROV2 training.}
\label{fig:success_bluerov}
\end{figure}

\begin{figure}[h!]
\centering
\includegraphics[width=0.92\linewidth]{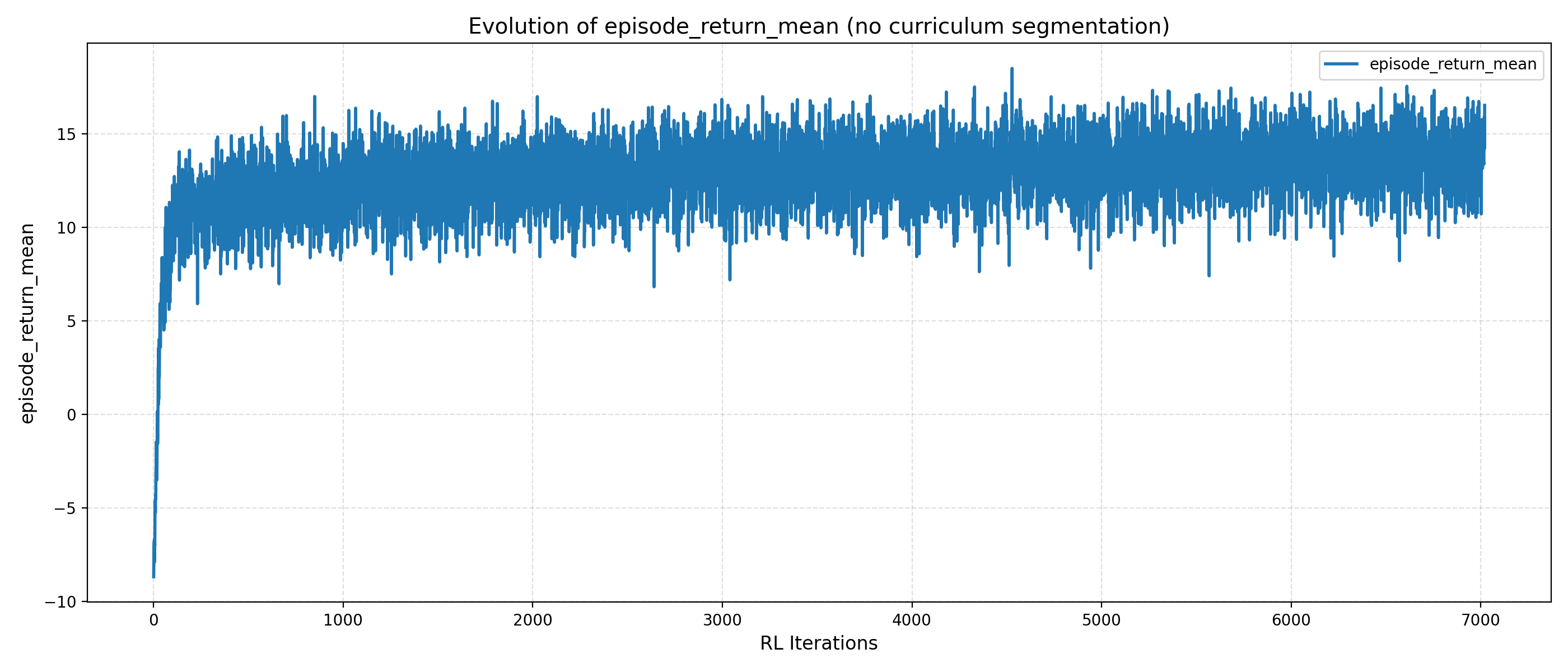}
\caption{Average return during training.}
\label{fig:return_bluerov}
\end{figure}

The associated curves are shown in Figures~\ref{fig:success_bluerov} and~\ref{fig:return_bluerov} and were computed over a total of 7020 training iterations.

\subsection{Evolution of the Success Rate}

The curve shown in Figure~\ref{fig:success_bluerov} exhibits a globally stable progression throughout training.
Statistical analysis indicates:

\begin{itemize}
    \item an average value of \textbf{0.734},
    \item a median of \textbf{0.745},
    \item a range from \textbf{0.0004} to \textbf{0.796}.
\end{itemize}

These results suggest that the agent quickly achieves and maintains a high success rate around 73--75\%. The very low minimum corresponds to the earliest iterations of training, before policy stabilisation.

The evolution of the success rate indicates that the agent progressively improves its ability to reach the exit without collision despite an environment containing up to ten obstacles. This stability is consistent with a well-converged and robust policy.

\subsection{Evolution of the Cumulative Reward}

The curve in Figure~\ref{fig:return_bluerov}, showing the evolution of the average reward per episode, follows a trend consistent with the improvement in the success rate. The computed statistics are:

\begin{itemize}
    \item mean: \textbf{12.77},
    \item median: \textbf{12.91},
    \item minimum: \textbf{-8.69},
    \item maximum: \textbf{18.50}.
\end{itemize}

Overall, the distribution remains centred around high values. The interquartile range (from 11.76 to 14.03) shows limited dispersion, a sign of a stable policy.

The coherence between the increase in average reward and the consolidation of the success rate confirms that the agent effectively learns to avoid collisions while optimising its trajectory toward the exit.

\subsection{Global Interpretation}

The joint analysis of the two metrics highlights an efficient, high-performing, and stable navigation policy:

\begin{itemize}
    \item a \textbf{robust success rate} around 74\% despite a complex environment,
    \item a \textbf{high average reward}, indicating a stable and optimised behaviour,
    \item a \textbf{low variability} across iterations, showing strong convergence.
\end{itemize}

In conclusion, the agent trained in an environment containing ten obstacles demonstrates solid performance in terms of episode success and trajectory optimisation. These results indicate reliable mastery of obstacle avoidance capabilities and generalisable behaviour across the configurations encountered during training.

\subsection{Comparison with the Reference Algorithm (DWA)}

To assess the robustness and effectiveness of the proposed PPO model, we conducted a systematic comparison with the \textsc{Dynamic Window Approach} (DWA). Both approaches were evaluated in a simulated environment containing ten obstacles, focusing on three operational metrics:
(i) success rate (reaching the target),  
(ii) collision rate,  
(iii) rate of exiting the operational area.

The results, obtained over 100 episodes in which the ten obstacles were randomly positioned identically for both navigation algorithms, are presented in Table~\ref{tab:comparison_dwa_bluerov}. They clearly show that the PPO model significantly outperforms DWA across most criteria. While DWA reaches the objective in only 8\,\% of episodes, the PPO model achieves a success rate of 55\,\%, nearly seven times higher. Similarly, the collision rate is drastically reduced with the learned policy (17\,\% versus 76\,\% for DWA), reflecting better obstacle anticipation and more stable decision-making.

Regarding zone exits, DWA records a lower rate (16\,\%) compared to PPO (28\,\%). This trend reflects the more exploratory behaviour of the RL model, which is capable of detouring around obstacles but sometimes at the cost of larger trajectory deviations.

Overall, these results demonstrate a clear superiority of the PPO model in terms of global performance, decision accuracy, and handling of complex navigation scenarios, thereby justifying its relevance for autonomous underwater missions.

\begin{table}[H]
\centering
\caption{Performance comparison between PPO and DWA over 100 episodes.}
\label{tab:comparison_dwa_bluerov}
\begin{tabular}{lccc}
\toprule
\textbf{Method} & \textbf{Success (\%)} & \textbf{Collisions (\%)} & \textbf{Out-of-area (\%)} \\
\midrule
DWA & 8 & 76 & 16 \\
PPO & 55 & 17 & 28 \\
\bottomrule
\end{tabular}
\end{table}

\subsection{Analysis of the Divergence Between Training and Test Success Rates}

We note that the average success rate observed during training (around 74\,\%) differs from the success rate obtained on the 100-episode comparative protocol (55\,\%). This divergence is expected and explained by several factors intrinsic to reinforcement learning and the experimental protocol.

First, the two metrics are not computed over identical episode distributions. The average training success rate is evaluated over more than 7000 iterations, covering a wide variety of obstacle configurations encountered throughout training, including early phases when the policy remains partially exploratory. In contrast, the 100-episode PPO–DWA benchmark uses a fixed protocol with ten obstacles randomly placed but identical for both algorithms. These scenarios, sometimes harder than the average training distribution, naturally reduce absolute performance.

Second, moving from an evolving policy during training to a fixed final policy introduces a shift. The agent evaluated in the benchmark no longer benefits from the continuous optimisation updates of PPO but adopts a stable deterministic behaviour that may be slightly less performant in some difficult cases. This reflects the effect of \textit{distribution shift} between the training experience and the fixed test distribution.

Finally, the small sample of 100 episodes introduces non-negligible statistical variability. In cluttered, highly stochastic environments, a policy with a true mean success rate around 70\,\% may reasonably achieve a lower observed rate on a small set of episodes, especially if these include a higher proportion of challenging configurations.

Despite this divergence, the PPO model vastly outperforms the DWA planner across all relevant metrics (55\,\% versus 8\,\% success), confirming that the learned policy generalises better than the deterministic method in unstructured environments. The difference between the two rates is therefore interpreted not as a limitation but as a typical characteristic of RL methods evaluated on test distributions more challenging than those encountered on average during training.

\subsection{Sea Experiments}

\begin{figure}[h!]
\centering
\includegraphics[width=0.8\linewidth]{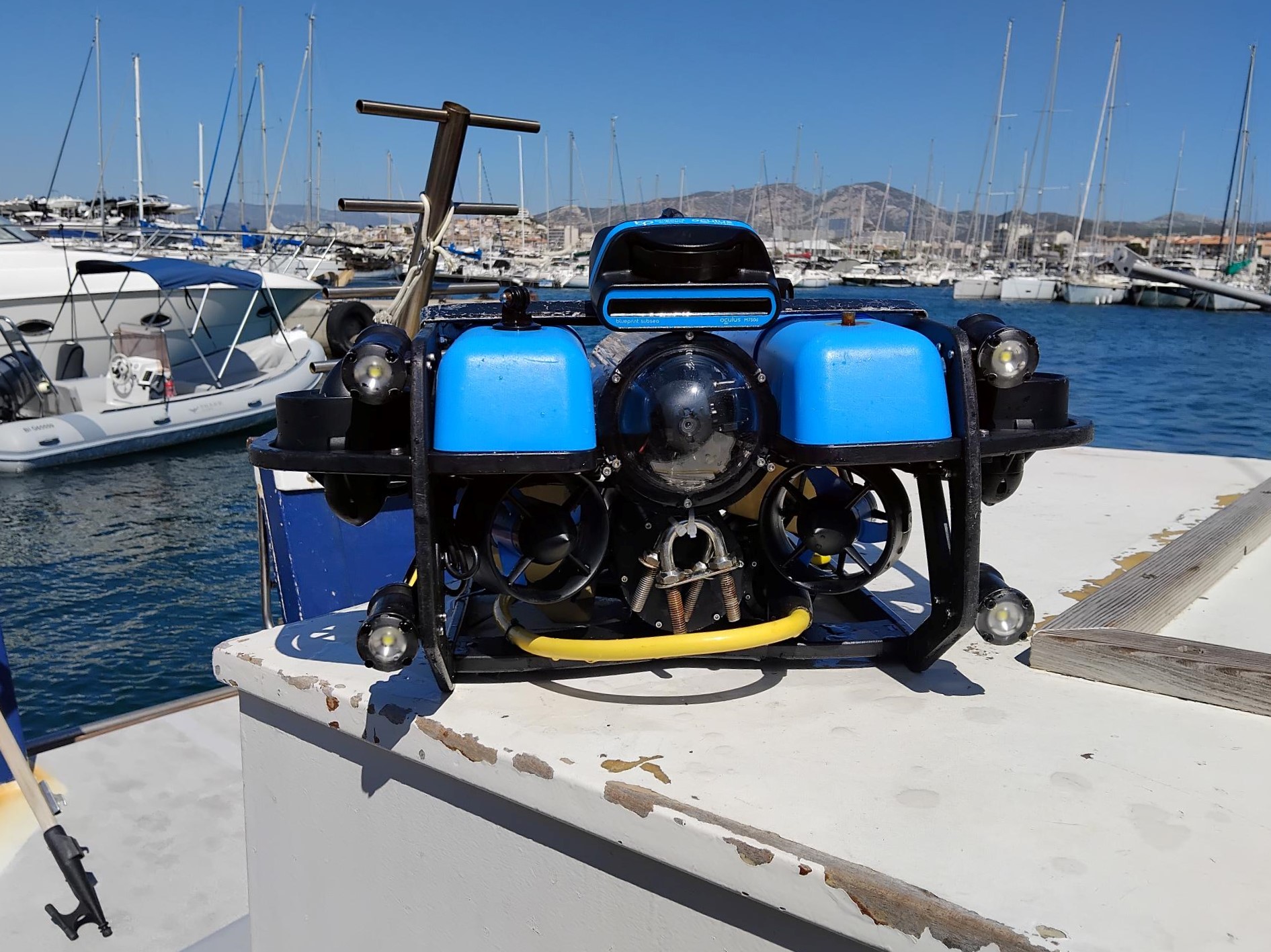}
\caption{Photograph of the BlueROV2 used during sea trials in the Pointe Rouge harbour (Marseille).}
\label{fig:bluerov2}
\end{figure}

The sea trials conducted in the Pointe Rouge harbour in Marseille consisted in testing the autonomous control of the RL agent on a \textit{BlueROV2 Heavy} and assessing its ability to avoid fixed obstacles materialised only within the 3D scene for hardware safety reasons.

\subsubsection{Hardware Architecture}

The \textit{BlueROV2} is equipped with inertial navigation sensors, an altimeter, and a forward-facing camera. Underwater localisation is ensured by an Ultra-Short Baseline (USBL) acoustic positioning system, providing sub-metric accuracy within a limited operational radius. A surface control station maintains communication with the robot through a waterproof Ethernet link, enabling bidirectional transmission of telemetry data and control commands.

\subsubsection{Software Architecture}

Control of the BlueROV2 relies on a software stack built on MAVLink, providing the interface between the onboard hardware and the RL decision module. The RL module continuously receives vehicle states (estimated position, orientation, obstacle distances) and sends back the optimal control actions. This real-time perception–action loop ensures the reactivity required for obstacle avoidance while maintaining stable behaviour.

\begin{figure}[h!]
\centering
\includegraphics[width=0.8\linewidth]{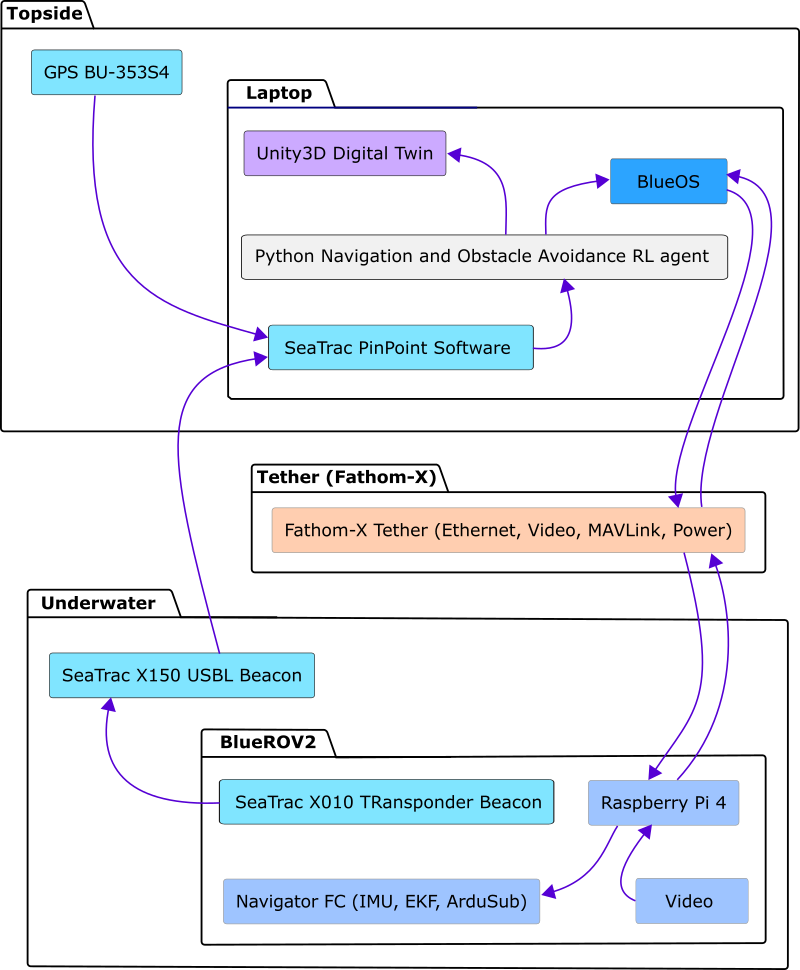}
\caption{Diagram showing the hardware and software components of the system used during sea trials.}
\label{fig:dig_twin}
\end{figure}

\subsubsection{Digital Twin and Visual Monitoring}

\begin{figure}[h!]
\centering
\includegraphics[width=0.8\linewidth]{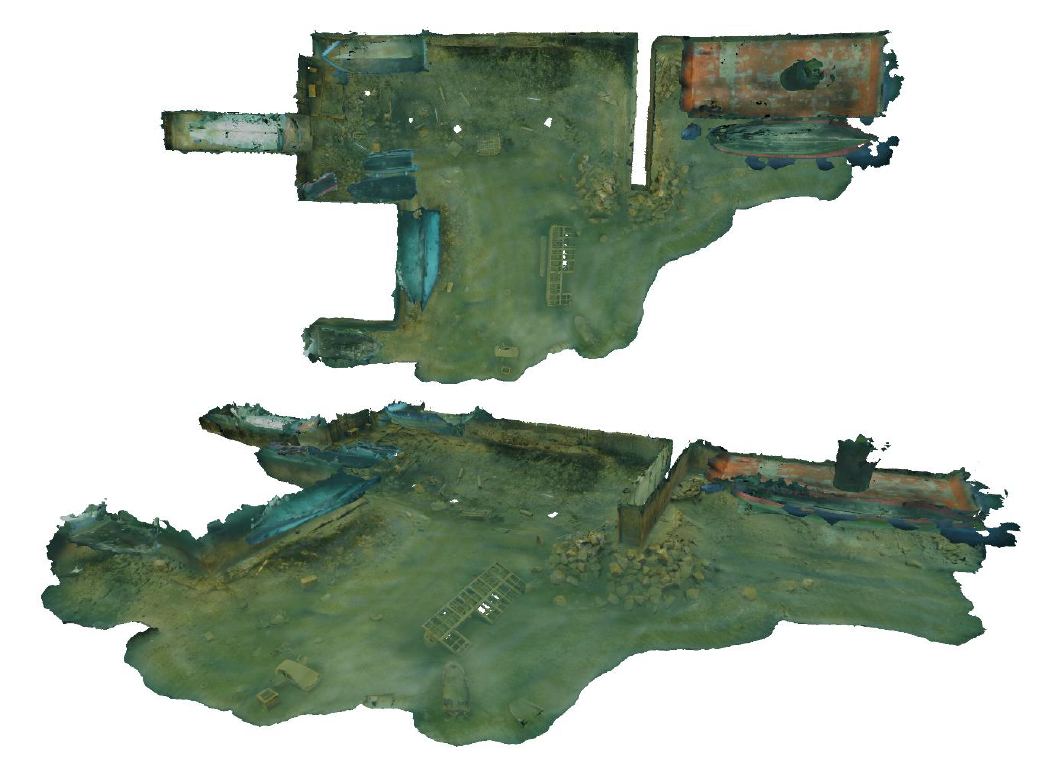}
\caption{Screenshot of the 3D model of the underwater test site reconstructed by photogrammetry.}
\label{fig:inpp}
\end{figure}

\begin{figure}[h!]
\centering
\includegraphics[width=0.8\linewidth]{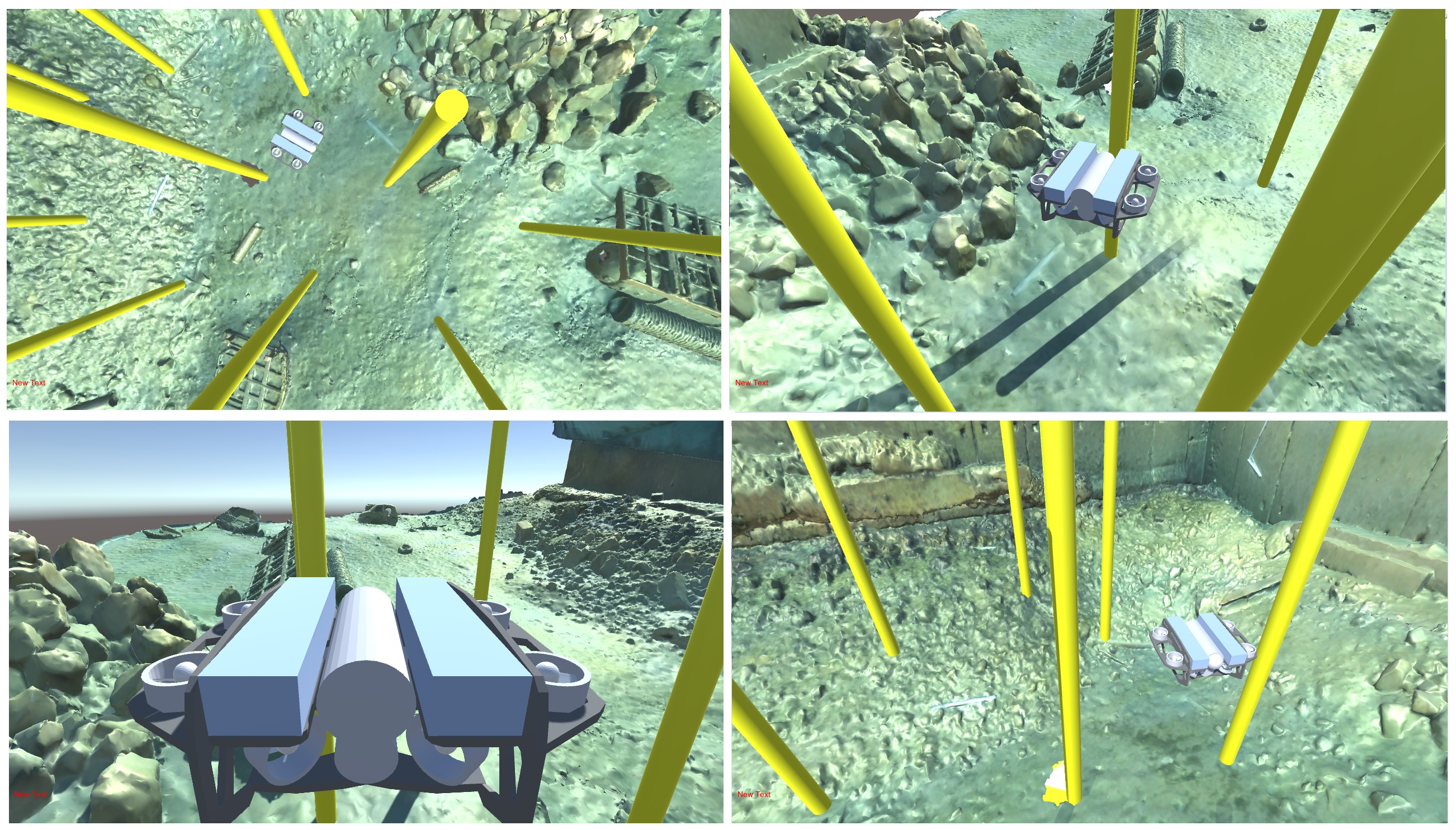}
\caption{Screenshot of the Unity3D environment showing the autonomous vehicle avatar in the simulated marine environment.}
\label{fig:unity3D}
\end{figure}

A digital twin of the test site was created from a 3D photogrammetric reconstruction and integrated into the Unity3D engine (version 6000.0.37f1). This twin is synchronised in real time with the position estimated by the acoustic system, allowing visualisation of the 3D avatar of the BlueROV2 and its virtual camera within the digital environment. This simulation–reality coupling provides visual supervision of the experiments and a direct comparison basis with the simulated trajectories.

\begin{figure}[h!]
\centering
\includegraphics[width=0.8\linewidth]{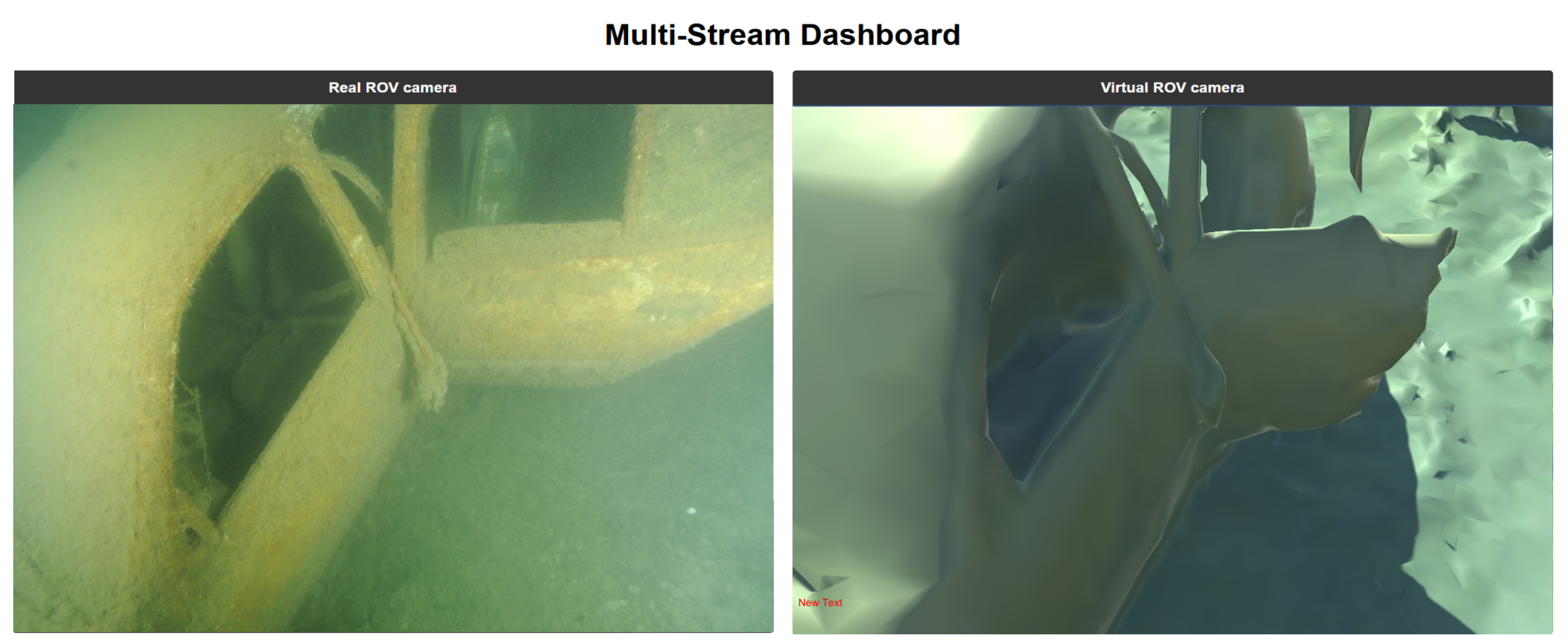}
\caption{Screenshots showing the synchronisation between the real BlueROV2 camera and that of its avatar within the 3D model of the test site. 
The real-world images were acquired under poor underwater visibility conditions, while the Unity-based 3D model of the site was deliberately downsampled from the original photogrammetry dataset to reduce rendering costs and ensure real-time performance of the simulation engine.}
\label{fig:virtu-real}
\end{figure}

\subsubsection{Results}

The navigation scenario consisted of reaching a fixed target while avoiding stationary obstacles defined exclusively within the 3D scene, ensuring zero physical risk during trials. Each experimental sequence validated the end-to-end functioning of the autonomous pipeline, from perception and localisation to RL decision-making and motor execution on the real vehicle.

The experiments conducted in real conditions confirmed the system’s ability to maintain behaviour consistent with that observed in simulation. The real \textit{BlueROV2} followed trajectories that were globally similar to those predicted in the virtual environment, and the avatar within the digital twin reproduced in real time the movements estimated by the acoustic system. This visual consistency, illustrated by the correspondence between real and virtual camera views (Fig.~\ref{fig:virtu-real}), demonstrates the effectiveness of the simulation–reality coupling and the synchronisation of data streams.

The deviations observed between the planned and real trajectories remained limited and were mainly due to noise inherent to acoustic positioning and to dynamic uncertainties of the vehicle in the harbour environment. Despite these sources of variability, the agent consistently pursued the target while maintaining a safe trajectory, thereby validating the perception–action loop on the real robot.

These initial results demonstrate the feasibility of deploying an RL policy trained in simulation to control a \textit{BlueROV2} in a secure real environment. They also show that the digital twin plays a crucial role by providing reliable visual supervision and an analysis support that facilitates the interpretation of observed behaviours. Overall, the trials validate the functional pipeline and confirm the relevance of this approach as a first step toward richer autonomous navigation incorporating, in future work, real perception sensors and more complex underwater environments.

\section{Discussion}

The results obtained in simulation show that the PPO-based autonomous navigation approach enables the learning of robust and effective policies in highly cluttered environments. By relying on a hybrid observation space combining target-oriented navigation cues, a local occupancy grid, and geometric information extracted via raycasting, the RL agent develops reactive behaviours that are difficult to achieve through manual tuning of a kinematic cost function. The direct comparison with the deterministic DWA algorithm reveals a clear advantage for RL in the most complex scenarios, confirming the relevance of learning-based methods for underwater robotics operating in irregular or densely constrained environments.

Validation in real-world conditions, made possible through the digital twin of the test site, provides complementary insights into system behaviour. The trajectories observed on the physical robot exhibit satisfactory qualitative consistency with those obtained in simulation, despite discrepancies attributable to acoustic positioning noise and dynamic uncertainties inherent to the harbour environment. These findings highlight the feasibility of sim-to-real transfer for RL policies while underscoring the importance of model discrepancies in underwater settings.

\subsection{Improving localisation through visual relocalisation in a photogrammetric model}

Sea trials emphasise the critical role of localisation in ensuring reliable autonomous control. At present, the reference trajectory of the \textit{BlueROV2} relies primarily on USBL acoustic positioning, whose accuracy and stability may vary depending on site configuration, and whose cost limits accessibility for small platforms. A particularly promising alternative is to exploit the image set used to construct the photogrammetric 3D model of the test site.

Rather than correlating real images with synthetic views rendered from the digital twin, the idea here is to establish a direct correspondence between the real-time image captured by the \textit{BlueROV2} camera and one of the source images used in the photogrammetry process, whose pose (position and orientation) is known with high accuracy. This principle of visual relocalisation within a database of \textit{pre-oriented} images could refine the estimation of the robot’s orientation—and potentially its local position—within the 3D model.

The prospects of such an approach extend beyond simple supervision. In port, industrial, or strategically sensitive sites that have been pre-modelled, visual alignment with a photogrammetric dataset could form the basis of a low-cost local underwater positioning system. By correlating images from the autonomous drone with those used to build the model, platforms lacking advanced acoustic sensors could achieve accurate localisation during inspection or surveillance missions.

This approach nevertheless presents significant challenges for real-time onboard use: managing large image datasets, ensuring descriptor robustness in turbid environments, coping with underwater lighting variability, and handling the computational cost of image-to-database matching. These limitations, however, represent valuable opportunities for future research.

\subsection{General perspectives}

Taken together, the results position this work as a key step toward underwater autonomy based on learning-driven policies. The consistency of transferred trajectories, the operational feasibility of the simulation–reality pipeline, and the superior performance over DWA in constrained environments constitute foundational milestones for more ambitious scenarios.

The identified future directions include:

(i) integrating real perception sensors (video, sonar) into the decision loop,

(ii) extending navigation to full three-dimensional control,

(iii) exploring \textit{safe RL} strategies to better regulate real-world behaviour,

(iv) investigating multi-agent policies for collaborative missions,

(v) and developing visual relocalisation approaches in photogrammetric models to reinforce vehicle localisation.

Ultimately, these avenues converge toward more robust, multimodal, and operational autonomy, in which learning-based policies interact seamlessly with realistic 3D models, diverse sensors, and complex underwater scenarios.

\section{Conclusions}

This work presented an initial validation of autonomous navigation for a \textit{BlueROV2} underwater vehicle based on deep reinforcement learning. The problem was formulated as a Markov decision process with a hybrid observation space combining target-oriented navigation information, a local environment representation through a virtual occupancy grid, and geometric perception via raycasting. Building on this formulation, a PPO policy was trained and evaluated within a 3D environment faithfully reproducing the physical constraints and obstacles of the operational domain.

The experiments show that the RL agent is capable of ensuring safe and efficient navigation in cluttered environments and that, in the most complex scenarios, it can surpass the DWA planner, widely recognised as one of the most common deterministic algorithms for obstacle avoidance. Validation on a physical \textit{BlueROV2}, enabled by a precise digital twin of the test site providing safe hardware conditions, confirms the transferability of the learned policy and establishes the feasibility of RL-based autonomous control.

This work represents a foundational step toward integrating learning-based policies into real underwater missions. By establishing a safe, reproducible framework closely linked to the physical robot, this milestone opens the path to a second phase involving real perception sensors (video, sonar) and trials in natural underwater environments featuring authentic obstacles and more diverse conditions. A particularly promising direction concerns improving relative positioning through visual relocalisation within pre-constructed photogrammetric models. Additional perspectives include extending navigation to 3D, investigating \textit{safe RL} methods to strengthen operational safety, and exploring multi-agent strategies for cooperative missions. Ultimately, these advances aim to bring RL-based autonomous navigation closer to real-world deployment.

\section*{Author Contributions}
Main manuscript writing, Z. Mari; design, development, and evaluation of the reinforcement learning agent, Z. Mari; 
review and proofreading, M.M. Nawaf; execution of sea trials and data acquisition, M.M. Nawaf and P. Drap; 
construction of the 3D site model and its integration into the simulation environment, M.M. Nawaf and P. Drap;
project coordination and scientific supervision, Z. Mari.

\renewcommand{\refname}{References}

\end{document}